\documentclass[letterpaper, 10 pt, conference]{ieeeconf}  % Comment this line out

\usepackage{floatrow}
\newfloatcommand{capbtabbox}{table}[][\FBwidth]

\usepackage{amssymb}
\usepackage{amsmath}
\usepackage{amsmath,mathtools}
\usepackage{times}
\usepackage{caption}
\usepackage[labelformat=simple]{subcaption}
\usepackage{graphicx}
\usepackage{graphics}
\usepackage{blkarray}
\usepackage{booktabs}
\usepackage{xcolor} 
\usepackage{bbold}
\usepackage{caption}
\usepackage{subcaption}
\usepackage{multirow}
\usepackage{comment}
\usepackage{epstopdf}
\usepackage[section]{placeins}
\usepackage{tikz}
\usepackage{todonotes}
\graphicspath{{./figures/}}
\usepackage{csquotes}
\usepackage{siunitx}
\usepackage{cite}
\usepackage{epigraph}
\usepackage{empheq}
\usepackage{comment}
\usepackage{cancel}% http://ctan.org/pkg/cancel
\usepackage{algorithm} 
\usepackage{algorithmic}  
\usepackage{mathtools}

\usepackage{balance}
\usepackage{dirtytalk}
\usepackage{pifont}% http://ctan.org/pkg/pifont

\usepackage[linesnumbered,ruled,vlined,algo2e,noend]{algorithm2e}

\SetInd{0.35em}{0.5em}

\usepackage{hyperref}

\SetKwRepeat{Do}{do}{while}

\makeatother

\newdimen\figrasterwd
\figrasterwd\textwidth

\DeclareCaptionLabelSeparator{periodspace}{.\quad}
\IEEEoverridecommandlockouts                              % This command is only
\newtheorem{example}{Example}

\makeatletter
\newcommand*{\rom}[1]{\expandafter\@slowromancap\romannumeral #1@}
\makeatother

\def\eqalignno#1{\let\\=\cr\displ@y \tabskip\@centering
  \halign to\displaywidth{\hfil$\@lign\displaystyle{##}$\tabskip\z@skip
    &$\@lign\displaystyle{{}##}$\hfil\tabskip\@centering
    &\llap{$\@lign##$}\tabskip\z@skip\crcr
    #1\crcr}}
\def\leqalignno#1{\let\\=\cr\displ@y \tabskip\@centering
  \halign to\displaywidth{\hfil$\@lign\displaystyle{##}$\tabskip\z@skip
    &$\@lign\displaystyle{{}##}$\hfil\tabskip\@centering
    &\kern-\displaywidth\rlap{$\@lign##$}\tabskip\displaywidth\crcr
    #1\crcr}}
\begin{document}

\thispagestyle{empty}
\twocolumn
\title{\LARGE \bf
 Address Behaviour Vulnerabilities \\ in the Next Generation of Autonomous Robots}

\author{Michele Colledanchise % <-this % stops a space
\thanks{The authors are with the Humaniods Sensing and Perception Lab, Istituto Italiano di Tecnologia. Genoa, Italy.
e-mail: \tt{ michele.colledanchise@iit.it}}}

\maketitle
\thispagestyle{empty}
\pagestyle{empty}
%%%%%%%%%%%%%%%%%%%%%%%%%%%%%%%%%%%%%%%%%%%%%%%%%%%%%%%%%%%%%%%

\begin{abstract}

Robots applications in our daily life increase at an unprecedented pace. As robots will soon operate \say{out in the wild}, we must identify the safety and security vulnerabilities they will face.
Robotics researchers and manufacturers focus their attention on new, cheaper, and more reliable applications. Still, they often disregard the operability in adversarial environments where a trusted or untrusted user can jeopardize or even alter the robot’s task.

In this paper, we identify a new paradigm of security threats in the next generation of robots. These threats fall beyond the known hardware or network-based ones, and we must find new solutions to address them. These new threats include malicious use of the robot's privileged access, tampering with the robot sensors system, and tricking the robot’s deliberation into harmful behaviors.
We provide a taxonomy of attacks that exploit these vulnerabilities with realistic examples, and we outline effective countermeasures to prevent better, detect, and mitigate them.

\end{abstract}
\section{Introduction}
\label{sec:introduction}

%While driving a car, people stopped to worry about XXX if the car behave as expected and responds correclty to commands. We somewhat trust that, provided a certain degree of maintenance, our car does not fail. This level of trust results from over two centuries of automotive development. However, according to the World Health Organization (WHO), vehicles sold in 80\% of all countries worldwide fail to meet basic safety standards.\footnote{\url{https://www.who.int/news-room/facts-in-pictures/detail/road-safety}}. 

When we withdraw money from an ATM or swipe a credit card in a gas station, we generally do not worry about the machine's behavior or the ATM's authenticity. However, ill-intentioned can easily alter a card reader by physically acting on it to send sensitive data to untrusted third parties~\cite{erickson2008hacking}. Such illicit activity, which yields an annual loss of \$24.26 billion in the USA only~\cite{smartmetric}, remains one of the financial industry's most difficult crimes to protect against~\cite{skimming}. This highlights to what extent we underestimate the security threats in our daily lives and how simple attacks can severely impact us.

%This highlights to what extend we underestimate the security threats in our daily life, and how simple attacks can have a severe impact.% on the society. 

The introduction of social robots in our societies will represent a fertile ground for a new security threat paradigm. These threats fall beyond the hardware and software ones and, without preventive measures, they can yield harmful robot behaviors. Examples include harmful physical interaction with the robot (e.g., tampering with road signs to attack self-driving cars), malicious use of privileged access of the robot (e.g., introduction of a factory defect in the food industry), and accidental use of the robot (e.g., a non-expert user designs a dangerous robot behavior). Conventional tools in cybersecurity alone may not quickly recognize and respond to attacks that exploit these vulnerabilities.

 %Additionally, customers and users place more value on cost and functionality of robots~\cite{mirjalili2008security}

We, as the robotics community, often give a lower priority to new security threats. On the one hand, robot manufacturers  prioritize development costs, speed to market, and customer satisfaction; on the other hand, robotics researchers focus on new robotics applications and new problems to solve~\cite{mirjalili2008security}.

\begin{figure}[t]
\centering
\begin{subfigure}[t]{\columnwidth}
\centering
\includegraphics[width=0.45\columnwidth, trim=0 3cm 0 1cm, clip]{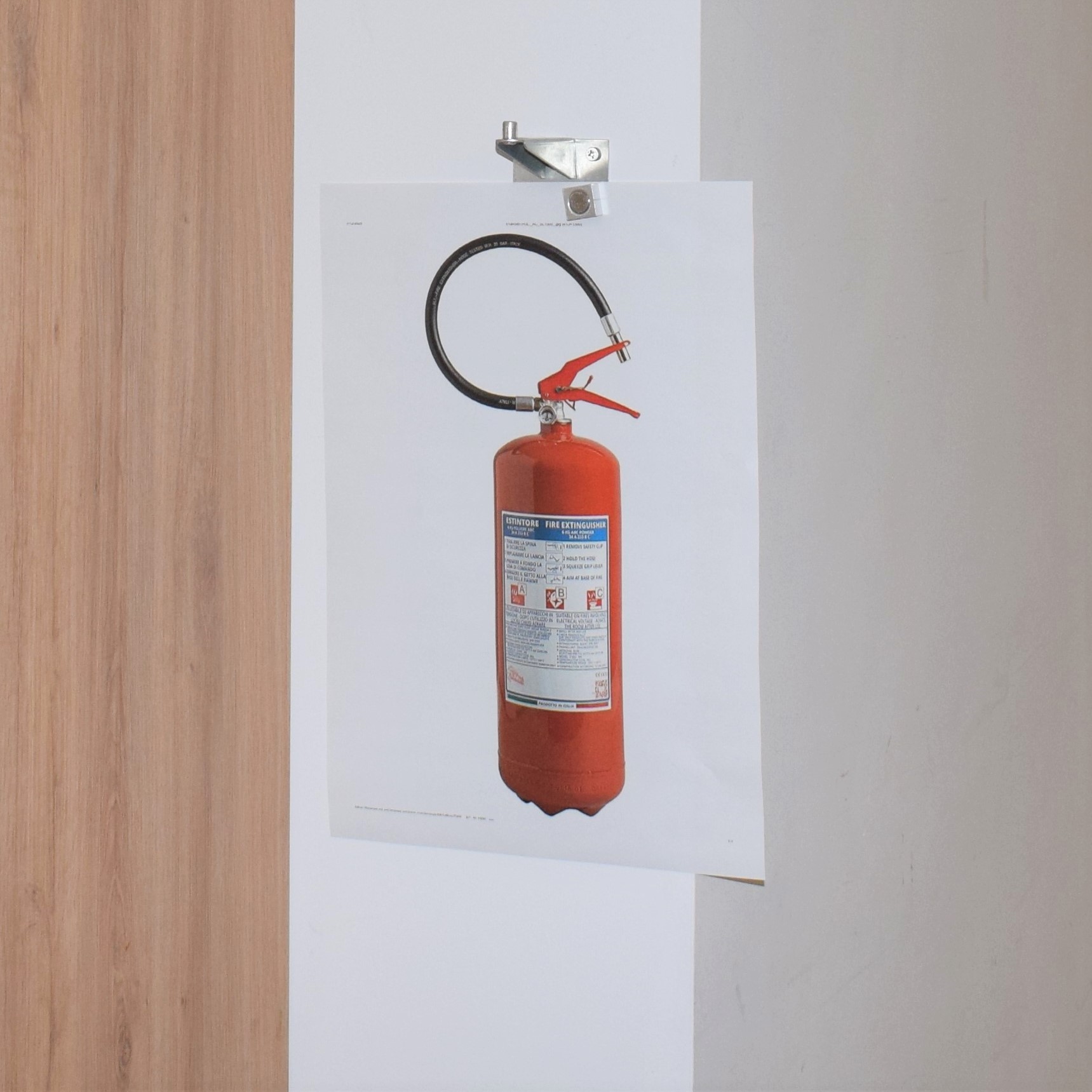}
%\end{subfigure}
\hfill
%\begin{subfigure}[t]{0.45\columnwidth}
%\centering
\includegraphics[width=0.45\columnwidth, trim=0 3cm 0 0, clip]{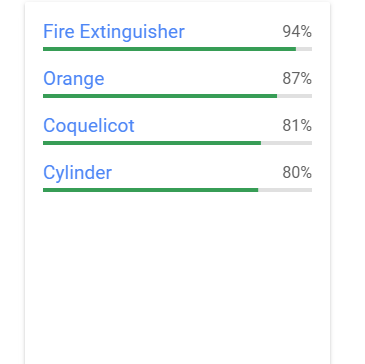}
\caption{
Printout of a fire extinguisher in place of a real one to deceive a surveillance robot (left) and the object detection results using Google Cloud API (right). }
\label{fig.intro.fire}

\end{subfigure}

\vspace*{1em}

\begin{subfigure}[b]{\columnwidth}
\centering
\includegraphics[width=0.45\columnwidth,trim=0 4cm 0 0cm, clip]{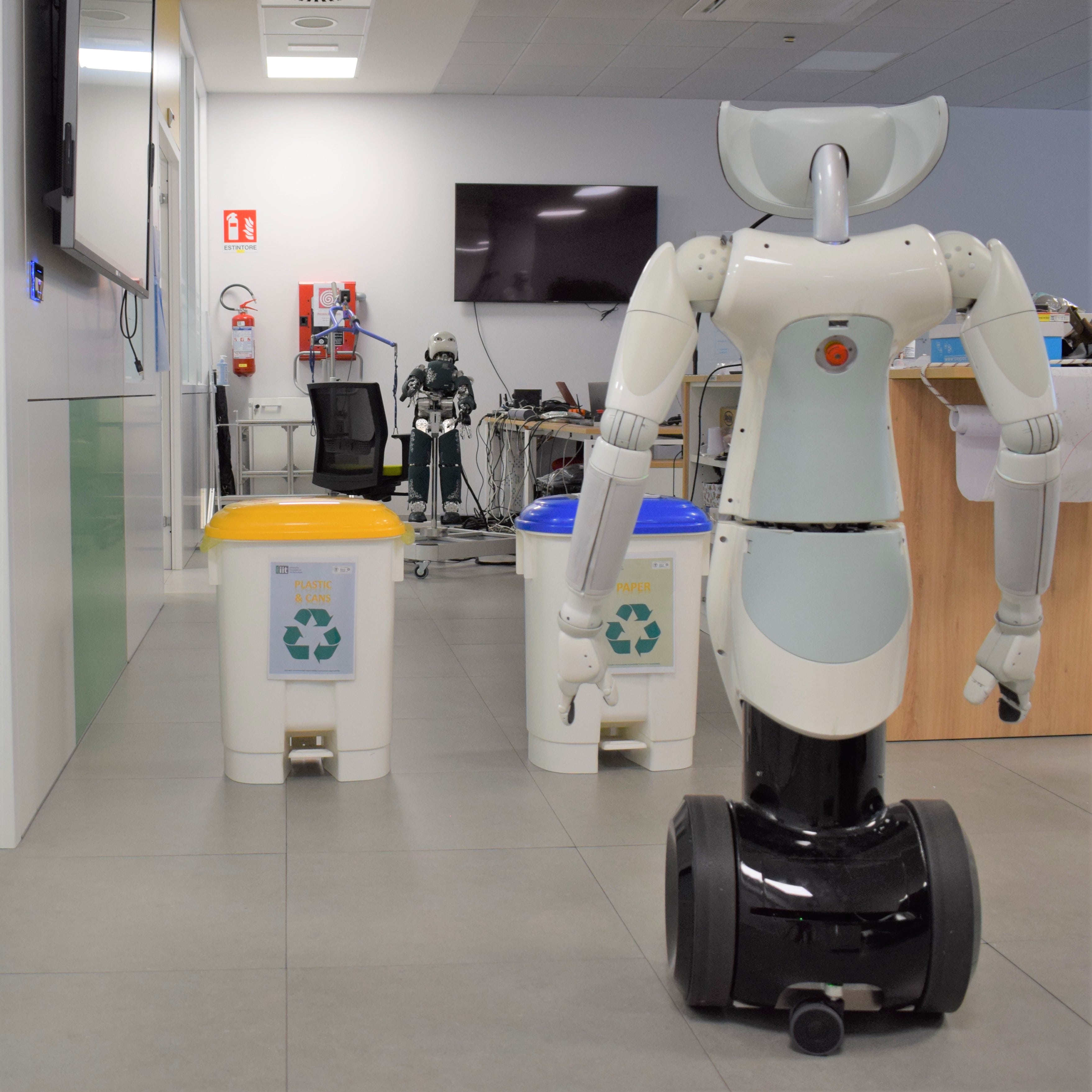}
\hfill
\includegraphics[width=0.45\columnwidth,trim=0 4cm 0 0cm, clip]{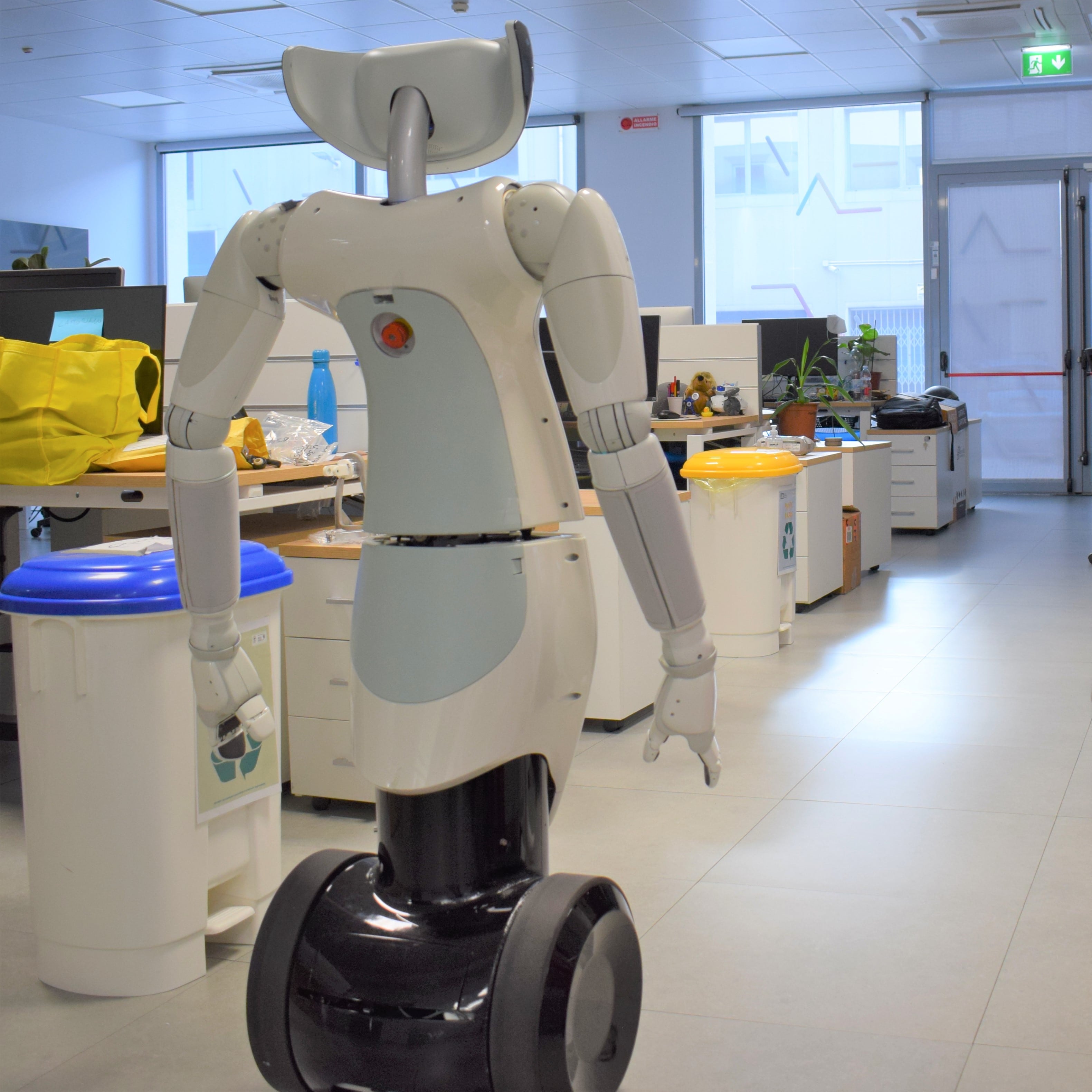}
\caption{Obstacles maliciously placed to hijack the robot.}
\label{fig.intro.r1}

\end{subfigure}
\caption{Examples of behavior threats carried out by non-expert users.}
\label{fig.intro}
\end{figure}
\vspace*{-1em}

We need countermeasures that combine cybersecurity and robots' deliberation tools to face this new paradigm of threats. Countermeasures incur certain costs from an economic and usability perspective, but they prevent much larger expenses.
Consider the example of an arsonist that deceives a surveillance robot by placing printouts of fire extinguishers in place of the real ones before setting the building on fire (as in Figure~\ref{fig.intro.fire}) or a felon that places obstacles on the robot's path to hijack it (as in Figure~\ref{fig.intro.r1}). Even using state-of-the-art navigation and object detection tools, as we did in the example, the robot is not only unable to perform its task correctly, but it is also unable to realize that an adversarial actor prevents the success of those tasks. 
The growing economic anxiety and the fear of robots displacing workers~\cite{hurt,scare} may motivate these attacks. Moreover, nation-states, criminal organizations, radical political or religious groups may sponsor them; and, more alarming, they do not require any prior knowledge in cybersecurity.
%\footnote{We used the NVIDIA's ISAAC SDK for the navigation and the Google Cloud API for the object detection}

This paper analyzes this new paradigm of robotics threats, provides a taxonomy of attacks and suggests possible effective countermeasures.
The analysis suggests a high risk of attacks as they will become easier to perpetrate and robots mass-produced.
We aim at drawing attention to this new security landscape and outline possible countermeasures.

\section{Existing Software and Hardware Security Threats in Robotics}
In this section, we outline the existing software and security issues in robotics.
Existing threats focus on intentional attacks that compromise the robot's hardware or software whereas we are interested in studying threats, both intentional and unintentional, that compromise the robotics deliberation, leaving intact the hardware and software.

\subsection{Hardware threats}
Once robots populate our homes and cities, attackers will easily have direct access to their sensors (e.g., cameras and laser scanners) and onboard hardware interfaces (e.g., USB ports and CPU's pins).
Moreover, robots will be mass-produced. Thus skilled individuals can reverse engineer robot components and possibly expose hardware vulnerabilities. This will reduce the expertise required to execute certain attacks.
Infected hardware can manipulate data being sent and received before the  Operating System (OS) level and therefore remain hard to detect~\cite{zaddach2013implementation}, as in the following example.

\begin{example}[Malicious Repair]
A domestic robot is sent to a local repair shop to fix its broken arm. In addition to the new arm, the repairer installs a hardware backdoor on the robot's camera so that they can steal sensitive data, as credit card numbers or personal footage used to blackmail the robot's owners.
\end{example}

Once an attacker takes a system's hardware, it becomes much harder to regain control of the system~\cite{montasari2020hardware,clark2017cybersecurity}.
Physical access represents one of the largest threats to robotics
and Internet of Things (IoT) devices~\cite{jouini2014classification,lofaro2016secure}.

\subsection{Software threats}

Robots rely on software libraries to perform their tasks, and an OS provides common services for these libraries. Thus robots
also have the same software threats as standard computers. Attackers have a long experience in malware (i.e., malicious software) development.
Moreover, the ability to mass upgrade the robots' software brings software threats to a much larger scale.

Robots will exchange information with each other to either communicate their intentions (e.g., the maneuver of autonomous cars) or to synchronize their interactions (e.g., handover of an object).
With the increasing reliance on interconnected robots, there will emerge new software threats that target this interconnection.
Robots can infect one another via their, possibly trusted, communication network.

\begin{example}[Domestic Hack]
A skilled cybercriminal hacks his domestic robots to install malware to each robot connected to it from the same brand. Each infected robot installs the malware to the other robots that it encounters. After a week, that malware infected all the robots of that brand, resulting in a financial loss for the manufacturer and the users. 
\end{example}

Fortunately, the long experience in developing antivirus programs and generic security software suites may tame the effects of software threats.

\newpage

\section{A new paradigm of security threats}
We foresee a change to the robotics threats landscape. This section outlines a new paradigm of security threats: \emph{behavior threats}. These threats exploit both the deliberation and the perception system's intrinsic vulnerabilities, tricking robots into tasks harmful for themselves and the humans surrounding them. We identify three types of attacks in this new paradigm: 1) the \emph{threats to robots} where attackers target the robots, 2) the \emph{threats through robots} where robots represent an attack vector, and 3) \emph{accidental threats} where the attack is performed without awareness (e.g., misuse or software bug).
Behavior threats can involve any software robot control architecture level, from the high-level deliberation to the low-level actuation control. 

The presence of a particular unresolved vulnerability implies that an attacker will eventually exploit it and the robot's behavior vulnerabilities are no exception.
For example, self-driving cars create the opportunity for attackers to cause crashes by performing adversarial behaviors. A stop
sign modified in specific ways, which humans
would easily recognize as still a stop sign, might
nevertheless, be misclassified by an AI system.
Moreover, as cybercriminals will probably employ AI systems to carry out labor-intensive attacks as spear phishing~\cite{brundage2018malicious}, we cannot exclude that they will eventually use AI systems to target robot's intelligence.

\subsection{Threats to robots}
Robot abuse exists since the first deployments in open areas of robots.
hitchBOT~\cite{smith2017hitchbot}, a robot that successfully hitchhiked across Canada, Germany, and the Netherlands, got decapitated in Philadelphia~\cite{herzfeld2015mourning} in 2015.
In Silicon Valley, a man beat to the ground a security robot~\cite{siliconattack}. In San Francisco, a group of people covered a robot in barbecue sauce and wrapped it in a tarp~\cite{bbqattack}. In Moscow, a man attacked a robot with a baseball bat while it pleaded for help~\cite{moskowattack}. In Osaka, three boys beat a humanoid in a shopping mall~\cite{osakaattack}.
However, these attacks remain mostly physical, and they are easy to detect, as the robot cannot continue any task. Future attacks will involve the robot's intelligence, and they will be harder to detect, as described in Section~\ref{sec:attacks} below.

%While human antagonism toward robots has different forms and motivations, it often resembles the ways that humans hurt each other. Robot abuse might stem from the tribal psychology of insiders and outsiders~\cite{Agnieszka}.

\subsection{Threats through robots}
Attacks through robots may leverage the privileged access of robots to a facility (such as hospitals or private offices) to perpetrate a crime. An attacker may physically modify a robot to implement malicious tasks, like introducing a wifi sniffer to steal sensitive data, a gun to kill a specific individual, or an explosive device to execute a terror attack.

\subsection{Accidental Threats}

Accidental threats refer to situations in which a human causes a dangerous robot behavior without awareness. They include programming(e.g., a bug in the behavior's code) and user (e.g., a misspelled request) errors.
These types of threats will probably become the most frequent ones as robots will have more inexperienced (or imprudent) users than attackers~\cite{nomura2016children, brvsvcic2015escaping}.

\clearpage

\section{Behavior Attacks Taxonomy}
\label{sec:attacks}
In this section, we outline the four categories for the attacks described in the previous section. We discriminate attacks based on intentionality and the level of expertise required. We do not consider software and hardware threats as they fall outside the scope of this paper.

\subsection{Intentional Physical Attacks}
Intentional physical attacks require physical access or interaction to the robot either during its use, maintenance, or even production. These attacks may be the most dangerous ones as they can provide additional robot capabilities as carrying an explosive or a gun. We now present two realistic intentional physical attack examples.
\begin{example}[Robot Kidnap]
\label{example.kidnap}
A domestic robot goes to a pharmacy to fetch drugs for its assisted human. On its way back, a couple of criminals places cardboard boxes maliciously in front of the robot, forcing it to change its paths to home. The criminals keep moving the boxes in front of the robot's new path in such a way that they force it to navigate through an alley, where they steal the robot undisturbed.
\end{example}

The attack example above does not exploit any hardware or software vulnerabilities of the robot and thus it results hard to detect algorithmically. 

\begin{example}[Explosive Warehouse Robot]
A warehouse robot fetches small packages and brings them directly to the customer. The robot has a LED battery indicator that turns red when the battery gets low.
A criminal glues an explosive device on top of the led indicator. Such a device, using a color sensor, activates the bomb when the red light is on. Once back in the warehouse, the robot battery gets low, making the device triggers the bomb.

\end{example}

The attack example above does not modify the robot's hardware or software, not even its behavior. Instead, it exploits a simple robot behavior, i.e., switching the color of a led, for malicious purposes.

\subsection{Unintentional Physical Attacks}
Unintentional physical attacks originate from na\"ive interactions with robots as asking to fetch a burning pot. But, more alarming, a na\"ive human behavior can represent an \emph{apparent threat} to others. Robots, especially those deployed in law enforcement, may have a dangerous reaction in front of an apparent threat to humans, as in the following example.

\begin{example}[Toy Guns]
A group of airsoft players exercises in an abandoned building. A bystander calls the police to report the trespassing. A police officer sends a reconnaissance drone to the building before deploying human officers. The drone scours the building until it sees a woman that shot at a man with a rifle. The drone then engages the Taser gun, neutralizing the innocent woman.

\end{example}

Unfortunately, these apparent threats remain profoundly hard to decode, and they deceive humans too. For example, three police officers shot to death a toy gun-carrying disabled man in Sweden~\cite{autismshoot2} and a police officer opened fire towards a man with autism deemed to hold his therapist hostage in California~\cite{autismshoot}.

\newpage
\subsection{Intentional Programming Attacks}

Intentional programming attacks involve the manipulation or exploitation of the robot's code to cause danger.
The introduction of malicious behaviors can be perpetrated at the production stage, from an inside cybercriminal, or during its use, from any generic felon.

The presence of several service robots in our cities opens the opportunity to execute a \say{Physical} Distributed Denial of Service (DDoS) attack, as in the following example.
\begin{example}[Physical DDoS]
\label{example.ddos}
A cybercriminal, using a bot, orders 1000 pizzas to different homes from 50 different fully-autonomous pizzerias. The locations of the target homes are near a hospital. During rush hour, the attack perpetrated let the ambulances (and other vehicles headed for the hospital) be stuck in a traffic jam.

\end{example}
The distributed nature of the above attackmakes it hard to detect.

Assistive and eldercare facilities already employ robots to help healthcare professionals in simple tasks~\cite{broekens2009assistive}. These robots will soon perform more complex tasks, including physical interaction with patients that will probably be programmed \emph{by demonstration}, where the robot observes the execution of a certain task from a human trainer and extracts relevant information for reproducing it~\cite{calinon2009robot}. 

State of the art in machine learning and robot's deliberation remains unable to discriminate if the task demonstrated hides malicious intents~\cite{brown2019machine}. A human trainer may program the robot to hurt patients, as in the example below.

\begin{example}[Malicious demonstration]
A nurse in an eldercare facility teaches the brand new robot how to lift non-self-sufficient patients. The nurse shows an operation that causes pain to the patients when lifted.
\end{example}

Episodes like the one above are not far from reality. A study suggested that nurses account for 86\% of murders perpetrated by healthcare professionals~\cite{yorker2006serial}.

\subsection{Unintentional Programming Attacks}
Unintentional programming attacks result from negligence, inattention, or lack of training while using or interacting with robots. They include programs with a correct syntax but a wrong semantic and programs with bugs that static program analysis tools cannot detect, as the insertion of a parameter with the wrong unit of measure (e.g., NASA lost a spacecraft due to a metric mistake~\cite{lloyd1999metric}).

\begin{example}[Autonomous Cars]
In Ogden, Riley county of Kansas, a woman asks her voice-controlled autonomous car to drive her sightless father to the county's capital, Manhattan, where his domestic robot is waiting for him.
Several hours later, the car brings the man to Manhattan in New York.
\end{example}

The unintentional attack above results from the overconfidence in autonomous systems' abilities. Nowadays, similar overconfidence causes the \emph{death by GPS} where drivers faithfully follow their navigation systems and disregard road signs, turning to the wrong road or drive away from populated areas and die before rescuers can reach them~\cite{lin2017understanding}.

%\begin{example}[Smart Home]
%\cite{zhang2020smart}
%\end{example}
%The impact of accidental attacks as the one above will be dramatic, as they affect thousand of robots around the globe.

\newpage

\section{Countermeasures}
In this section, we outline possible technological countermeasures to behavior threats. Political and social measures fall outside of the scope of this paper. We divide countermeasure into three groups prevention, detection, and mitigation, providing realistic examples.

\subsection{Prevention}
 Following the  fundamental principle of modern health care 
\say{prevention is better than cure}~\cite{gandy2011perspective}, prevention represents probably the most important countermeasure to behavior threats.

The development of less vulnerable robots' software and hardware helps in reducing behavior vulnerability. However, behavior threats emphasize the need for prevention measures like the ones below.

\paragraph*{Sustainable and agile robot's code development}
The design and maintenance of a secure architecture, from a software and hardware point of view, represent a well-known problem for generic embedded systems and IoT~\cite{boot1993security}. On the one hand, the robots' interconnection will allow quick software, and hence security, updates, but on the other hand represents an attack vector, both intentional and unintentional. We may want to limit the rate of these software updates~\cite{simoens2018internet}. Moreover, we must strive to make software and hardware development \emph{sustainable}.

Model-driven methods and tools enable sustainable software development for robotics~\cite{stampfer2018contributions}.
We must identify the structures and abstractions that
help design, integrate, and update complex robotic software systems.~\cite{lotz2018managing}

Software integration and update will very likely remain a challenge for the development of safe robots and, in general, for future robotic developments, as recognized by the European Commission~\cite{bogue2014future}.

%the software development is mostly based on community efforts and it is not coordinated by key stakeholders; (b) robotic technologies are characterized by a high variability that makes reuse of software a challenging practice; and (c) robotics developers are usually not specifically trained in software engineering. 

\paragraph*{Hierarchical behaviors}
Robot's behavior customization represents an attack vector. Both na\"ive  and malicious users may design robots behaviors that are dangerous for themselves and others. 

\begin{figure}[h]
\centering
\includegraphics[width=0.95\columnwidth]{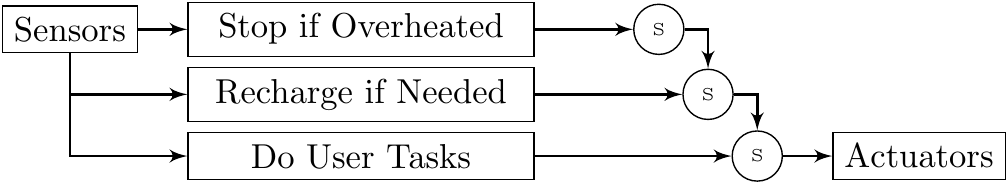}
\caption{Subsumption Architecture composed by three behaviors. The behaviors Stop if Overheated subsumes the behaviors Recharge if Needed, which subsumes the behaviors Do
User Tasks.}

\label{fig.subs}
\end{figure}

The robot's behavior architecture must support hierarchies. The attacks' prevention and mitigation behaviors must remain at a higher level in the hierarchy. Users should not be allowed to modify or even visualize such behaviors. Solutions as Behavior Trees~\cite{colledanchise2018behavior}, Subsumption Architecture~\cite{brooks1986robust}, or Hierarchical Task Networks~\cite{ghallab2016automated} will help in making that hierarchy simple. Figure~\ref{fig.subs} shows an example of a robot behavior hierarchy encoded with a Subsumption Architecture. Such behavior avoids that the robot overheats for too long, regardless of the user task. However, the behavior designer must ensure preemptable task execution.

\newpage
\paragraph*{Robust behaviors}
The example in Figure~\ref{fig.subs} suggests that the robots must operate correctly in different environmental conditions. Following the IEEE standard glossary of software engineering, we define this property as \emph{robustness}.
Behavior attackers may leverage on the inability of a robot to operate in a different setting than the ones expected and in the inability to overcome
task failures, hence robots must operate properly in case of such failures.

\paragraph*{Verified behaviors}
We need to identify languages to define precisely non-ambiguous task specifications, constraints, and methods to verify them.
We may employ formal methods used to specify and verify software systems to ensure the correctness of a robot's behavior~\cite{cortesi2013static,luckcuck2019formal}.

We could employ formal methods to design safe and secure behaviors, which prevent intentional and unintentional programming attacks, respectively. A verification framework could perform a static analysis on a new robot behavior, also defined by the user, to detect possible flaws in the behavior's logic and prevent programming attacks.

The \emph{safety} aspects in robotics are somewhat well defined,  in industrial robots' applications, from a human-robot interaction perspective~\cite{michalos2015design,shukla2016application,villani2018survey,heyer2010human}.
Unfortunately, there is less attention on the \emph{security} aspects of robot behavior development~\cite{clark2017cybersecurity}. However, extensions on formal methods should provide sufficient evidence for specific security certification~ \cite{murino2019resilience,pulina2010abstraction}.

\begin{figure}[b]
\centering
\includegraphics[width=0.9\columnwidth]{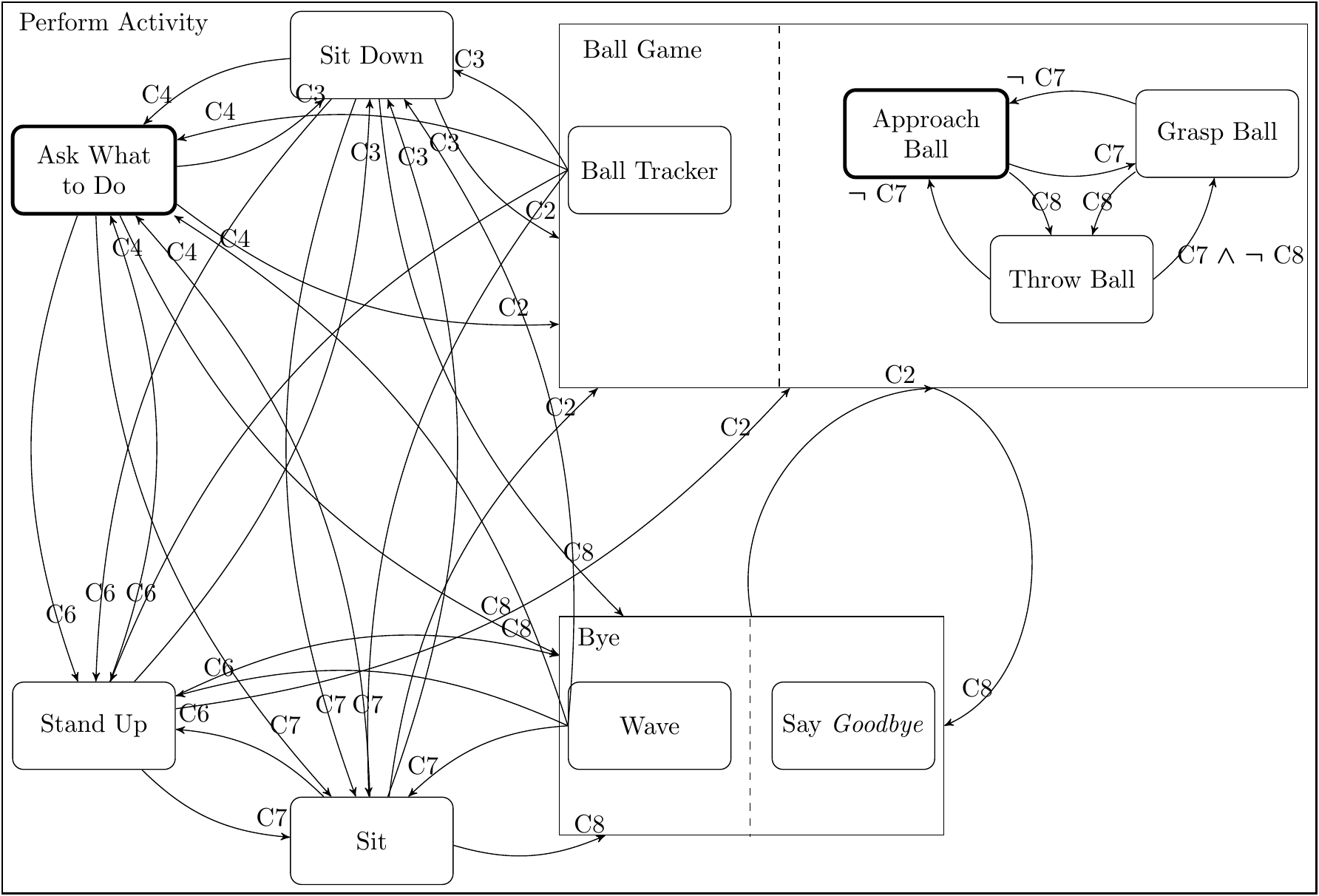}
\caption{Example of a non-readable behavior. Adapted from~\cite{colledanchise2018behavior}.}
\label{fig.readable.fsm}
\end{figure}

\paragraph*{Human readable behaviors} 
Borrowing the definition of readability from software programming, we can define readable behaviors essentially the ones easy to read and understand~\cite{kraeling2013embedded}. Consider a robot that performs different activities reactively according to the user inputs taken from the literature~\cite{colledanchise2018behavior}. Figure~\ref{fig.readable.fsm} depicts the behavior described as a Hierarchial Finite State Machine (FSM). A user may find such behavior descriptions hard to understand and edit.
Encoding the robot behavior with a readable representation lowers the likelihood of programming errors~\cite{paxton2018evaluating} from both the end-user and the manufacturer, which mitigates the unintentional threats. 
Human readable behaviors relate with the concept of explainable AI and improve the end-users trust that the robots make good decisions and the manufacturer accounts for possible damages~\cite{parmar2018robert}.

\newpage
\subsection{Detection}
The ability to detect attacks becomes fundamental to mitigate them. Intrusion detection and security challenges fall outside the scope of this paper. 

The anomaly detection tasks results especially challenging when
they must treat data from black boxes (e.g, Neural Networks) and without a formal definition of nominal settings~\cite{gupta2018open}. We suggest the following detection mechanism. 
%There are not many methods to
%eliminate false alarms beyond creating methods with better
%sensitivity and specificity

\paragraph*{Quality of Service Monitoring}
Robot behavior must be flexible enough to easily accommodate user requests and execute user's programs in different conditions. 
In both social and industrial robotics, a challenge is to guarantee the system's correctness while fulfilling Quality-of-Service (QoS) requirements. In this context, several QoS monitors exists~\cite{bouasker2020qos}.
We could define an explicit representation of the robot's behavior as non-functional properties, as done in~\cite{brugali2018model}, and exploit QoS monitors to detect an attack, as one of such qualities will possibly degrade.

\begin{figure}[b]
\centering
\includegraphics[width=0.9\columnwidth]{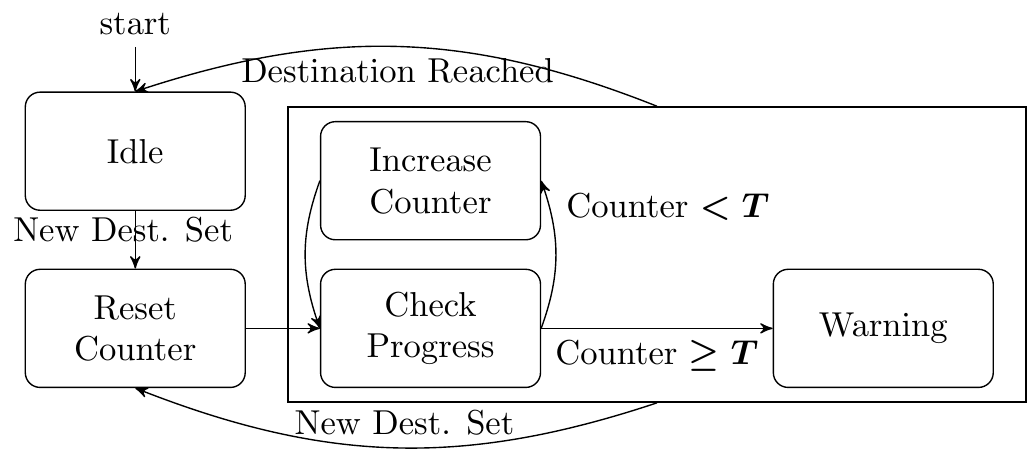}
\caption{Example of a runtime monitor that detects the attack of Example~\ref{example.kidnap}.}
\label{fig.runtime}
\end{figure}

\paragraph*{Redundant Software} 
We could employ formal methods tools to synthesize monitors that detect unexpected changes or anomalies in the robot's behavior at runtime.
Runtime monitoring models a software verification technique that involves observing the internal operations of a software system and its interactions with its environment (the robot and its surrounding in our context) to determine whether the system violates a given
specification~\cite{cassar2017survey}. We could use these runtime monitors to define anomaly detection systems. For example, the runtime monitor in Figure~\ref{fig.runtime} monitors the progress of a navigation task. Whenever a new destination is set, the (Hierarchical) FSM checks if the robot reaches the destination within a given time threshold $T$. Such a monitor could detect the attack described in Example~\ref{example.kidnap}.

\paragraph*{Learning Techniques}

Anomalies remain inevitably unpredictable in form and often expensive to miss. We can formulate anomaly detection problems as classification tasks~\cite{koller2009probabilistic}. A supervised or semi-supervised machine learning framework may train anomaly detection models, such as neural networks or support vector machines, from a dataset containing labeled instances of nominal and failed behaviors traces.
Unsupervised learning requires less input from a human expert, but they perform very poorly compared to supervised or semi-supervised learning~\cite{chandola2009anomaly,omar2013machine,pang2020deep}.

%We could employ reinforcement learning techniques to automatically synthesize detection models from a set of labeled, or partially labeled, anomalies as done in the literature~\cite{}.

Not all anomalies represent
failures. A robot can behave anomalously without failing. Conservative detection mechanisms cause false alarms that reduce the trust in robots~\cite{gupta2018open}.

\newpage
\subsection{Mitigation}
Once the robot detects an attack, it must actuate a mitigation strategy. 
A na\"ive mitigation strategy could be to disable the robot's actuators. However, this strategy may lead to dangerous behaviors, as in an autonomous car driving on the highway or in a drone flying over a crowd.
If the robot could not prevent such attacks, it means that the robot is in a situation for which its behavior is not intended. Here, the robot's deliberation capabilities play an important role since, if the robot fails to recover from an attack, it requires external intervention, which may be expensive.

\paragraph*{Task Planning}
Informally, task planning represents a set of tools to choose and organize actions to achieve a purpose.
As the robot's environment is, in general, uncertain and dynamic, we must employ task planning techniques that can operate in such a context.
Partially Observable Markov Decision Processes (POMDPs) provide a precise and elegant mathematical formulation for task planning in uncertain and dynamic environments and they found applications in various robotic tasks. However, solving POMDPs exactly remains computationally intractable~\cite{russell2016artificial, ingrand2017deliberation, ghallab2016automated}.

Unfortunately, a framework that solves long-horizon task planning in uncertain and dynamic environments lies far beyond the state of the art~\cite{russell2016artificial, kaelbling2013integrated, ghallab2016automated}.
Early works rely on the \emph{most-likely-state} approximation where they identify the most likely physical state of the robot, and then act assuming that the robot is on that state~\cite{Platt-RSS-10}. However, those methods do not plan for actions to reducing uncertainty and remain impractical for our purpose.
We may employ existing PDDL-based planning frameworks\cite{hadfield2015modular,kaelbling2013integrated} and their extensions in terms of replanning and goal reconsideration \cite{levihn2013foresight}; or task planners that employ a search algorithm in belief space~\cite{kim2019pomhdp, ruiken2016reconfigurable,colledanchise2020act}.

The AI and the robotic communities identified different key open challenges in this context\cite{ghallab2014actor,ghallab2016automated,ahmadi2020risk,wongpiromsarn2012receding}. In particular, task planning techniques often underestimates the reasoning
and deliberation needed for acting~\cite{ghallab2014actor} advocating for autonomous systems that \say{reacts to events and extends, updates, and repairs its
plan based on its perception}.

\paragraph*{Operators deployed from a Control Center}
Autonomous failure recovery will probably remain unfeasible for all possible attacks. Once the robot fails all its mitigation strategies, we could employ a rescue squad composed of humans or other robots to fix the robot on the spot or take it for maintenance. 

The rescue squad may be called by the robot itself, humans, or other robots nearby. However, this opens additional challenges in terms of false alarms and hoax calls to the rescue squads, with potentially dangerous consequences~\cite{crowder2018false}.

The mitigation strategies above rely on the fundamental assumption that the actions are \emph{reversible} i.e., their effects on the robot and the environment can be undone by a finite set of actions~\cite{kaelbling2011hierarchical} that the robot, or the rescue squad, can perform.
\newpage

\section{Concluding Remarks}
\label{sec:conclusions}
In this paper, we advocated a focus on a new paradigm of robotics threats that will sit alongside the software and hardware ones:  the \emph{behavior threats}. As robots will soon be part of our society and will play a major role in several countries' economic development, we must foresee the new security landscapes and shape effective countermeasures.
 
Our economy will strongly rely on social and industrial robots; attacks \emph{to} or \emph{through} them will have a severe economic impact.
The spread of a biological virus in our society resembles a software virus's spread in a computer network. A recent study compared the economic impacts of a cyber-attack to robots, both social and industrial, to that of a pandemic or a natural disaster~\cite{clark2017cybersecurity}. Cybercriminals will use robots to automate existing labor-intensive cyberattacks as spear phishing and create a new paradigm of attacks that exploit human vulnerabilities as the impersonation of domestic robots or robot kidnap and ransom demand. These attacks will probably target people from vulnerable socio-economic groups as lonely older people fond with their caretaker robot.

Our society will rely on social and industrial robots too. The impact of the malicious use of robots opens new scenarios of criminals activities as the recruitment of members and supporters of terrorist organizations, or the implementation of a \say{physical} DDoS, as in Example~\ref{example.ddos}. Scientists showed evidence on robot's persuasive effects on humans' sense of trust and willingness to
interact with them that result more effective than other technologies~\cite{rossi2018socially,staffa2016recommender}.
 Additionally, robot's social and human-like behaviors during generic tasks execution \cite{di2014role}, collaborative activities~\cite{broquere2014attentional}, and in response to humans'  emotions bring humans to interact with them as with other human beings. Scientists employed humanoids robots, with human-like behavior, as childrens' lecturers for educational scopes \cite{zaraki2018novel}, and for neuro-cognitive rehabilitation programs for disabled people \cite{rossi2018psychometric}. % and in the health-care domain \cite{cavallo2011multidisciplinary}.
Moreover, interactions with robots may affect human ethics, especially in terms of children's moral development. A recent study suggested that cruelty towards robots may make humans more insensitive towards other humans too~\cite{coghlan2019could}.

As with all new technologies, there will exist different threats when we deploy the next generation of robots. The consequences of robot behavior attacks must alarm us since,  unlike threats on software deployments that most often compromise data or business continuity, they can put human lives in danger.  
There will be an extended period during which it will be impossible to foresee and prevent all these attacks.
It is imperative that we, as members of the robotics community, strive to shorten that period.

\section{Acknowledgment}

This work was carried out in the context of the SCOPE project, which has received funding from the European Union's Horizon 2020 research and innovation programme under grant agreement No 732410, in the form of financial support to third parties of the RobMoSys project.

\balance

\bibliographystyle{IEEEtran}
\bibliography{refs}

\begin{thebibliography}{10}
\bibitem{autismshoot}
``What happened next to the man with autism whose aide was shot by police.''
\end{thebibliography}
\end{document}